\documentclass[conference,a4paper]{IEEEtran}
\IEEEoverridecommandlockouts

\usepackage[hidelinks]{hyperref}
\usepackage[cmex10]{amsmath}
\usepackage{amssymb,amsfonts}
\usepackage{dblfloatfix}

\usepackage[ruled,vlined]{algorithm2e}
\usepackage{graphicx}
\graphicspath{{Figures/PDF/}{Figures/PNG/}}

\usepackage{booktabs}
\usepackage{siunitx}
\usepackage[numbers,compress]{natbib}
\usepackage{texnames}
\usepackage{bm,bbm}
\usepackage{xspace}
\usepackage{tabu}
\usepackage{amssymb}
\newlength\savewidth
\newcommand{\tablestyle}[2]{\setlength{\tabcolsep}{#1}\renewcommand{\arraystretch}{#2}\centering\footnotesize}

\usepackage{multirow}
\usepackage[inline]{enumitem}

\begin{document}

\title{Unlocking Multi-Spectral Data for Multi-Modal Models with Guided Inputs and Chain-of-Thought Reasoning}

\author{	\IEEEauthorblockN{Dahun Kim}
	\IEEEauthorblockA{\textit{Google DeepMind}}
	\and
	\IEEEauthorblockN{Ganesh Satish Mallya}
	\IEEEauthorblockA{\textit{Google DeepMind}}
	\and
	\IEEEauthorblockN{Anelia Angelova}
	\IEEEauthorblockA{\textit{Google DeepMind}}
}

\maketitle
\begin{abstract}
Multi-spectral imagery is a valuable input signal for Remote Sensing applications, such as land-use and land-cover classification and environmental monitoring.
However, generalist Large Multi-modal Models (LMMs) are typically trained on RGB images, limiting their applicability to the RGB domain.
At the same time, training multi-spectral multi-modal models is expensive and produces uniquely specialized models.
To address this, we propose a novel training-free approach that introduces multi-spectral data within the inference pipeline of standard RGB-only LMMs, allowing large gains in performance.
Our approach leverages the LMMs' understanding of the visual space by adapting non-RGB inputs to that space and injecting domain-specific information and Chain-of-Thought reasoning as instructions.
We demonstrate this with the Gemini 2.5 model and observe strong Zero-Shot performance gains on popular Remote Sensing benchmarks. These results highlight the potential for geospatial professionals to leverage powerful generalist models for specialized sensor inputs, benefiting from rich reasoning capabilities grounded in specialized data.
\end{abstract}

\section{Introduction}

Large Multi-modal Models (LMM) serve as powerful generalist learners, exhibiting visual understanding capabilities that generalize well beyond training data. While Remote Sensing (RS) applications can benefit from these task-agnostic models, most generalist LMMs are trained solely on RGB inputs, rendering them unsuitable for multi-spectral data (e.g., Sentinel-2, Landsat). Multi-spectral bands are crucial for capturing material properties invisible to standard sensors, and combining them yields more accurate responses than RGB-only approaches, which suffer from inherent blind spots.

\begin{figure} [t]
   \centering
     \includegraphics[width=\linewidth]{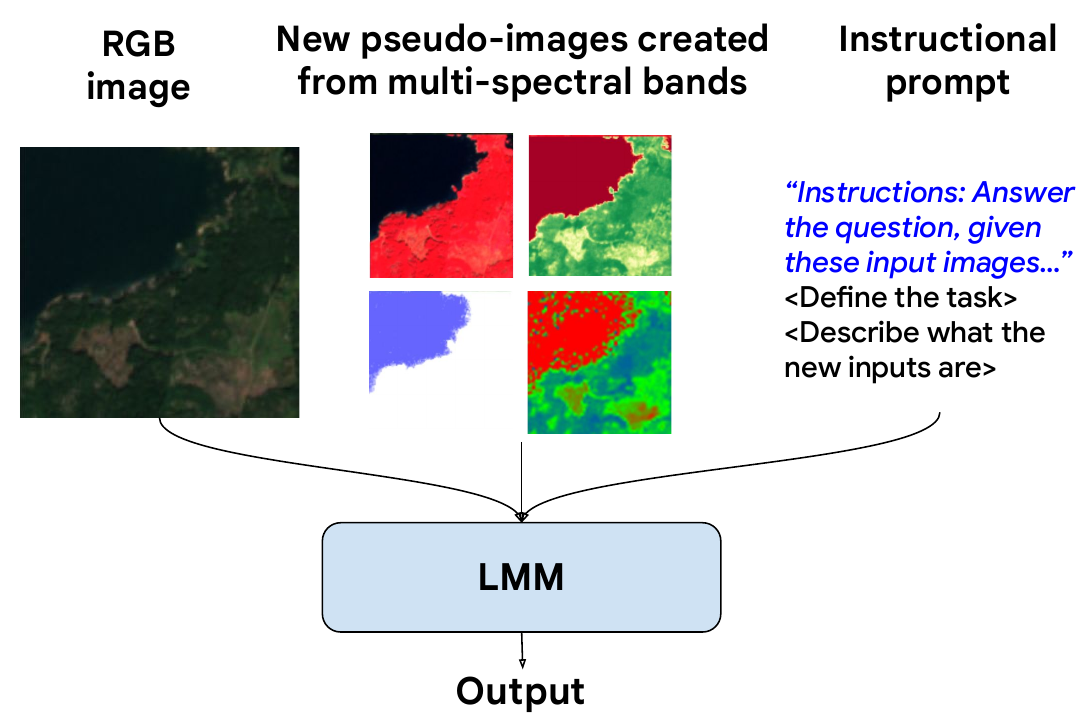}
     \vspace{-5mm}
   \caption{A generalist Large Multi-Modal Model (LMM), intended for RGB-only inputs, can be adapted, when queried Zero-Shot and without any training, to understand new and unfamiliar multi-spectral inputs. This improves the LMM's already strong performance on these tasks and extends its applicability to more Remote Sensing tasks which often rely on  multi-spectral inputs.
   }
   \vspace{-3mm}
   \label{fig:teaser}
\end{figure}

A common solution is training domain-specific foundational models~\cite{satmae2022,Prithvi-100M-preprint,SpectralGPT,HyperSigma} adapted for modalities like NIR (Near Infrared) and SWIR (Short-Wave Infrared). However, replicating large models on sensor-specific data is computationally expensive and data-intensive. 
Furthermore, rapid sensor evolution makes specialized models fragile. Changes in specifications, such as new spectral bands or higher resolutions, often render models incompatible, forcing expensive retraining. Therefore, developing adaptable frameworks that ensure LMM resilience to changing sensor inputs without re-training is critical.

We propose to re-purpose generalist multi-modal models to interpret unseen multi-spectral modalities by leveraging their inherent visual understanding and instruction-following capabilities. Our training-free approach represents new inputs as pseudo-images contextualized by extensive text descriptions (Figure~\ref{fig:teaser}). 
For example, using Sentinel-2 L2A 12-band data, we generate false-color and pseudo-color images, including band combinations (NIR, SWIR) and visualizations of indices like NDWI (Normalized Difference Water Index) and NDMI (Normalized Difference Moisture Index). While these pseudo-images remain within the model's visual space, we provide interpretative prompts detailing their creation, specifying included bands and physical meanings (e.g., ``indicative of moisture" or ``represents green vegetation"). This effectively grounds the generalist LMM's visual representations in the specific application domain.
We prepared a Colab tutorial\footnote{\scriptsize{github.com/google-gemini/cookbook/blob/main/examples/multi\_spectral\_remote\_sensing.ipynb}} which demonstrates the pipeline for generating multi-spectral pseudo-inputs and performing basic zero-shot inference. The advanced inference strategies presented in this paper are extensions beyond this basic tutorial.

We demonstrate this framework using the Gemini 2.5 model~\cite{Gemini2.5}, observing substantial improvements in an entirely Zero-Shot setting. Notably, our approach achieves new state-of-the-art (SOTA) performance, confirming that generalist LMMs can be effectively enabled to interpret data from novel input sensors without the need for additional training or fine-tuning. 
Our primary contribution is a streamlined mechanism that allows RGB-trained Multi-Modal models to seamlessly process unfamiliar multi-spectral data. This approach effectively unlocks new capabilities for generic LMMs, extending their applicability to many Remote Sensing tasks that were previously reliant on specialized adaptations.

\section{Previous work}

Multi-modal foundational models are widely deployed for Remote Sensing (RS)~\cite{FoundSurvey}, typically via domain-specific training. Prominent examples such as RemoteCLIP~\cite{RemoteCLIP}, SkyCLIP~\cite{skyscript}, and RS-CLIP~\cite{RSCLIP} are commonly used but are generally restricted to the RGB domain. Other strategies involve fine-tuning pre-existing models or pre-training specifically on RS data~\cite{Barzilai2025ARF, SatlasPretrain, SkySense, EarthPT, RemoteCLIP, skyscript, RSCLIP, FoundSurvey, kuckreja2023geochat, Scale-MAE, GFM, Billion}, occasionally employing lightweight adaptations~\cite{LoRA-NIR}. Despite their utility, these approaches predominantly work in the RGB-only domain.

Building foundational models for multi-spectral or multi-sensor data~\cite{SkySense, satmae2022, Prithvi-100M-preprint, SpectralSpatial, hong2022spectralformer, MaskedHyperspectral, SpectralGPT, CROMA, OmniSat, MMEarth} addresses this gap but presents significant challenges due to input diversity. SatMAE~\cite{satmae2022} extends pre-training to multi-spectral imagery, while the Prithvi series~\cite{Prithvi-100M-preprint} explicitly defines input channels for Blue, Green, Red, Narrow NIR, SWIR 1, and SWIR 2. SkySense~\cite{SkySense} integrates RGB, multi-spectral, and temporal inputs. In~\cite{Interband}, authors propose training by separating bands based on resolution: M1 (60x60m bands 1, 9), M2 (20x20m bands 5, 6, 7, 8A, 11, 12), and M3 (10x10m RGB and NIR). Furthermore, joint-training across Sentinel-2 and Sentinel-1 inputs demonstrates robust performance improvements~\cite{linial2025Enhancing}.
Specialized models also target hyper-spectral capabilities. SpectralGPT~\cite{SpectralGPT} proposes a spectral foundational model using a 3D Generalized Transformer, while HyperSIGMA~\cite{HyperSigma} utilizes sparse attention to optimize spectral-spatial feature extraction. 
While more comprehensive in their input modalities, these foundational models are not flexible in terms of adding new input channels. Furthermore training multi- or hyper-spectral models is a very resource-intensive task.
Another type of approach is to learn semantically meaningful embeddings from large number of Remote Sensing inputs~\cite{MOSAIKS}, ~\cite{SatlasPretrain},
~\cite{EarthPT}, 
~\cite{s2vec},
~\cite{AlphaEarth}. These features can integrate various input sensors, but are appropriate for clustering or further  fine-tuning. 
In contrast, our work used generalist RGB-based LMMs 
that are not specific to Remote Sensing. 
We demonstrate that such models can accommodate unfamiliar multi-spectral inputs in a Zero-Shot fashion without any model training or adaptation. This work expands upon a preliminary, unpublished technical report~\cite{mallya2025zero}, extending it with advanced reasoning strategies.

\section{Method}

\subsection{Zero-shot inference with unseen multi-spectral inputs} 
\label{sec:sensors} 
\begin{figure}
    \centering
    \includegraphics[width=0.081\textwidth,angle=90]{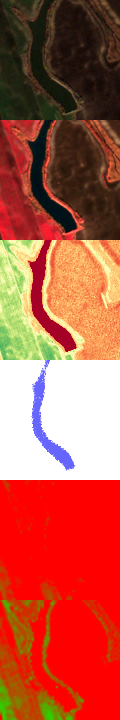}
    \vspace{-6mm}
    \caption{
    Examples of the six input modalities derived from the multi-spectral BigEarthNet dataset.
    Column 1: RGB image, Column 2: False Color composite, Column 3: NDVI (vegetation index), Column 4: NDWI (water index), Column 5 and 6: NDMI (moisture index) maps.
    }
    \label{fig:ben_examples}
    \vspace{-2mm}
\end{figure}

We adapt generalist RGB-trained Multi-Modal models to unseen multi-spectral modalities in a Zero-Shot fashion, avoiding the need for retraining. We transform multi-spectral inputs (e.g., Sentinel-2 L2A) into pseudo-images with visualizations of non-RGB bands (NIR, SWIR) and indices (NDVI, NDWI, NDMI), and contextualize them with detailed prompts mapping visual features to physical properties (Figure~\ref{fig:teaser}). 
The model leverages these ''visualized" spectral data points alongside the original RGB image (Figure~\ref{fig:ben_examples}).

\subsection{Prompting strategies for Zero-shot inference}

Due to the scarcity of paired multi-sensor data for training, we employ zero-shot prompting methods to adapt generalist LMMs. 
While all proposed methods outperform the RGB-only baseline, we find that Chain-of-Thought reasoning is the most effective with clear gains over simpler prompting schemes.

\subsubsection{Informative prompting} 
\label{sec:generic}
We supply the model with the newly generated pseudo-images alongside the original RGB input. To bridge the modality gap, we append a detailed instructional prompt describing the creation of each image, specifically specifying the constituent spectral bands and their physical significance.
This approach allows the LMM to process multi-spectral inputs through its standard image encoder while the text prompt provides critical context.
For example, we explicitly list the Sentinel-2 band composition within the prompt: 
\begin{itemize*}[label=]
    \item B02: Blue (10m);
    \item B03: Green (10m);
    \item B04: Red (10m);
    \item B05: Red Edge (704.1nm, 20m);
    \item B06: Red Edge (740.5nm, 20m);
    \item B07: Red Edge (782.8nm, 20m);
    \item B08: NIR (10m);
    \item B8A: Narrow NIR (20m);
    \item B01: Coastal Aerosol (60m);
    \item B09: Water Vapor (60m);
    \item B11: SWIR (1613.7nm, 20m);
    \item B12: SWIR (2202.4nm, 20m),
\end{itemize*}
noting the \textit{central wavelength} and/or \textit{spatial resolution} in the parentheses.

Leveraging this spectral context, we further describe the specific input composites:
\begin{itemize*}[label=]
    \item \textbf{RGB:} Composited from B04, B03, B02;
    \item \textbf{False Color:} Composited from B08, B04, B03;
    \item \textbf{NDVI:} Normalized Difference Vegetation Index (Red-Yellow-Green map) using B08, B04;
    \item \textbf{NDWI:} Normalized Difference Water Index (range -0.8 to 0.8) using B03, B08 with linear colormap [(1, 1, 1) to (0, 0, 1)];
    \item \textbf{NDMI-1:} Moisture Index using B8A, B11 with linear colormap [(1, 0, 0) to (0, 0, 1)];
    \item \textbf{NDMI-2:} Moisture Index using B8A, B12 with linear colormap [(1, 0, 0) to (0, 0, 1)].
\end{itemize*}

\subsubsection{Zero-Shot Inference with Vocabulary Expansion} 
\label{sec:expansion}

We enhance prompt quality through vocabulary expansion, specifically targeting common failure modes where class names are linguistically ambiguous (e.g., `Annual' vs. `Permanent Crop'). By analyzing these ambiguities, we refine class definitions to include key visual attributes and context, effectively teaching the model to distinguish targets at inference time. 

Used in conjunction with the input descriptions in Section~\ref{sec:generic}, we append category-specific guides to the prompt. For instance: ``(1) Agro-forestry: Trees mixed with crops/pasture; (2) Arable land: Cultivated land showing geometric patterns." This provides the necessary semantic constraints for accurate remote sensing classification.

\subsubsection{Zero-Shot Chain-of-Thought Multi-Step Reasoning}
\label{sec:cot}

We propose a Chain-of-Thought (CoT) approach utilizing a `Propose-and-Verify' style reasoning. Building on the seminal work of Wei et al.~\cite{cot}, we demonstrate its effectiveness in the context of our Zero-Shot multi-spectral inference.

In this advanced reasoning framework, the model is guided to propose hypotheses and verify them through a targeted step-by-step process. Specifically, we instruct the model to adhere to the following logic:
\begin{itemize*}[label=]
\item \textbf{1. Propose:} The model lists multiple potential classes.
\item \textbf{2. Verify:} The model re-investigates the images to find evidence for each proposal.
\item \textbf{3. Conclude:} The model selects final class(es) and provides the output.
\end{itemize*}

As an example, the prompt may incorporate the following Chain-of-Thought (CoT) instructions:
\textit{Review the information below and use the following three-step reasoning process: Analyze, Confirm, and Synthesize.
Step 1: Propose: Based on an initial examination of the 6 images, list 2-3 potential classes from the list below that might fit the scene. Briefly explain your initial reasoning for each proposal.
Step 2: Verify:
For each class you proposed, systematically verify it by finding specific visual evidence in the images. State whether you can confirm or deny the class. You MUST cite which image(s) and what features support your verification.
Step 3: Conclude:
Based on your verification, state your final conclusion. 
}
This instruction block is appended to the multi-spectral input prompt described in Section~\ref{sec:generic}.

This approach leverages the powerful visual representations of generalist LMMs, interpreting new visual inputs in a manner specialized for Remote Sensing applications. Crucially, this utilizes the model in a purely Zero-Shot capacity, requiring no additional fine-tuning. Consequently, this method achieves the highest Zero-Shot inference performance, demonstrating significant gains over previous techniques and strong baselines.

\subsection{Implementation details}
We generate five pseudo-images alongside the standard RGB input: 1) a false color composite using bands B08, B04, and B03; 2) an NDVI image (B08, B04); 3) an NDWI image (B03, B08); and 4–5) two NDMI images utilizing bands B8A with B11 and B12, respectively. Color maps are applied where necessary. For the true color and false color composites, we normalize each band to $[0, 1]$ and scale the values to $[0, 255]$ before stacking them into final RGB images.

\section{Experiments}

We evaluate the proposed approach on the three prompting strategies: 
\textbf{Baseline} is Zero-Shot inference with the new multi-spectral data and the informative prompt from Section~\ref{sec:generic}. 
\textbf{Expansion} uses the prompt described in Section \ref{sec:expansion}.
\textbf{CoT} (Chain-of-Thought) is with our Chain-of-Thought reasoning proposed in Section~\ref{sec:cot}.  
We report Zero-Shot results, demonstrating the performance of the model at inference-only, i.e. without training or fine-tuning, 
The same Gemini 2.5 model is used for inference, for either RGB or multi-spectral inputs.
Our evaluation protocols and model querying are aligned with recent work~\cite{Zhang2024GoodAt} which benchmarked GPT-4V and several other methods in the Zero-Shot setting, as well.
We compare to the state-of-the-art Zero Shot performance including for multi-spectral inputs, and to the RGB-only performance as a baseline. While we acknowledge that RGB-only is disadvantaged as it receives fewer inputs, we report it in order to illustrate the improvements achieved here by incorporating multi-spectral imagery, without any additional training or inference costs overhead. 

\subsection{Zero-Shot Results on BigEarthNet}

We first evaluate the performance of the proposed approach on BigEarthNet~\cite{BigEarthNet}, which is a land cover classification dataset which provides 12
multi-spectral bands.
BigEarthNet is a multi-label dataset, which means that more than one label are considered correct per example. For that reason the main metrics to evaluate this dataset is the F1 metrics, which is also consistent with performance reporting in the literature~\cite{Zhang2024GoodAt}. We also report Precision and Recall values to gain understanding of these components and their contribution to the final F1 metrics.
As  established in the literature~\cite{BigEarthNet19Cl,Zhang2024GoodAt}, we report the results on BigEarthNet with 19 classes, as the original 43 classes are ambiguous, and have overlapping semantic meaning.
Since the dataset is multi-label, we prompt the model specifying that more than one class is possible as an output. We do not limit the number of output classes for the model.

\begin{table}[h!]
\caption{
Zero-Shot results on BigEarthNet land cover classification (multi-label).
We compare against SOTA methods. The integration of multi-spectral and our advanced reasoning method achieves a gain of \textbf{+0.11} F1.
    }
    \vspace{-1mm}
    \centering
\tablestyle{5pt}{1.1}
\begin{tabular}{l|c|cc}
\toprule
Model & F1 & Precision & Recall \\
\midrule
GPT-4V~\cite{Zhang2024GoodAt}   &0.38  &0.49 &0.43 \\
Qwen-VL-Chat~\cite{Zhang2024GoodAt} &0.40  &0.57 &0.39  \\
InstructBLIP-FLAN-T5-xxl~\cite{Zhang2024GoodAt} &0.02  &0.41 &0.01  \\
InstructBLIP-Vicuna-13b~\cite{Zhang2024GoodAt}  &0.01  &0.01 &0.06\\
LLaVA-v1.5~\cite{Zhang2024GoodAt}  &0.39  &0.27 &0.83 \\
\midrule
Llama3-MS-CLIP~\cite{Llama3-MS-CLIP} (Multi-Spectral) &0.369 &0.267 &0.778\\
\midrule
Gemini RGB  &0.414  &0.542 &0.335 \\
\midrule
Gemini Multi-Spectral Baseline (Ours) &0.453  &0.566 & 0.377 \\
Gemini Multi-Spectral Expansion (Ours) &0.507  &0.535 &0.481\\
Gemini Multi-Spectral CoT (Ours) &\textbf{0.523}  &0.552 &0.502 \\
\bottomrule
\end{tabular}

    \label{tab:zs_comp19}
\end{table}

Table~\ref{tab:zs_comp19} shows the large performance improvements of the proposed multi-spectral approach,
over the already very strong performance of the RGB-only Gemini model. This is realized without needing any additional training or adaptation. We note the particularly large gains of the proposed Chain-of-Thought prompting.
We also note the very large gains achieved over previous state-of-the-art (SOTA) Zero-Shot methods, including Multi-Spectral ones (Table~\ref{tab:zs_comp19} top). As seen, Gemini 2.5 already has strong RGB-only performance, compared to other powerful models e.g. GPT-4V. Its performance is further improved by the proposed multi-spectral approach.
We further note that some prior works might trade-off precision for recall e.g. LLaVA-v1.5 has very high recall but low precision. Qwen-VL-Chat has slightly higher precision but low recall. 
Our approach outperforms both by large margin in F1 score, and also has a better balance of precision and recall.

\subsection{Zero-Shot Results on EuroSat}

In Table~\ref{tab:zs_eurosat} we evaluate the performance on the popular EuroSat benchmark~\cite{helber2019eurosat} for land use multi-class classification.
Its main performance metric is accuracy. 
The proposed approach obtains very large accuracy gains on this benchmark. Similarly here, we observe that the CoT method has the highest gains, while the other proposed approaches are also highly performant, outperforming the SOTA.
We also include results of ZLaP
which uses Inductive Zero-Shot inference, meaning it is allowed to use the dataset examples, but without labels. As seen, our approach, outperforms other powerful SOTA approaches, including Multi-Spectral SOTA ones. 

\textbf{Comparison to fine-tuned models}. While 
fully-trained domain-specific models can achieve high accuracy by learning explicit class-label associations via fine-tuning, they require costly data collection and per-dataset training. In contrast, our approach offers a generalizable, training-free alternative that adapts immediately to new tasks without model modification.

\begin{table}[]
\caption{Zero-Shot classification on EuroSat benchmark compared to SOTA (top section). Chain-of-Thought (CoT) with multi-spectral demonstrates impressive performance. 
    }
    \vspace{-1mm}
    \centering
    \tablestyle{10pt}{1.1}
    \begin{tabular}{l|c}
    \toprule
    Model & Accuracy, Top 1 (\%) \\
     \midrule
   CLIP~\cite{radford2021clip} (CLIP-VIT L/14-336px) &59.6 \\
    CLIP~\cite{radford2021clip} (CLIP-VIT L/14) &59.9 \\
    ZLaP~\cite{ZLaP} (Inductive inference) &63.2 \\
    \midrule
    GRAFT~\cite{GRAFT} (Multi-Spectral) &63.8 \\
    Llama3-MS-CLIP~\cite{Llama3-MS-CLIP} (Multi-Spectral) &67.8 \\
     \midrule
Gemini RGB & 66.3 \\    
\midrule
  Gemini Multi-Spectral (Basline) &69.1 \\
  Gemini Multi-Spectral Expansion (Ours) &70.7 \\
  Gemini Multi-Spectral CoT (Ours) &\textbf{72.7} \\
     \bottomrule
    \end{tabular}
    \vspace{-2mm}
    \label{tab:zs_eurosat}
\end{table}

\textbf{Ablations.} Table~\ref{tab:big_earth_ablations} details the impact of input modalities and prompting strategies. Baseline prompting shows a clear benefit from spectral data, where adding all multi-spectral bands improves F1 score. Vocabulary Expansion (\ref{sec:expansion}) further boosts the performance. Chain-of-Thought (\ref{sec:cot}) reasoning proves to be the most effective strategy.

We further analyze the contribution of the informative prompt descriptions (\ref{sec:generic}) within the CoT framework. Removing the specific descriptions of the pseudo-images or the spectral band definitions results in a performance drop. This indicates that the specific domain grounding provided by the instructional prompts is important for the model to effectively leverage the multi-spectral data. 
Finally, the combination of CoT with All Multi-Spectral inputs and full descriptive prompting yields the highest performance (0.534 F1).

\begin{figure}[t]
  \centering
  \includegraphics[width=\linewidth]{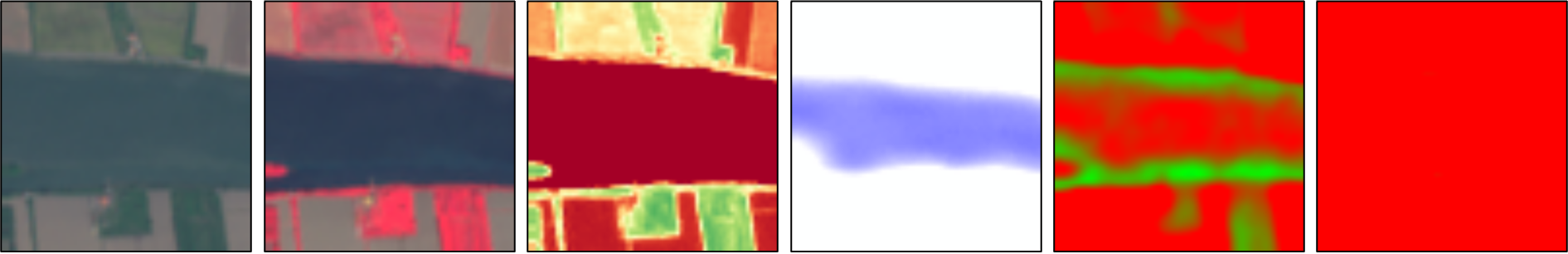}
  \includegraphics[width=\linewidth]{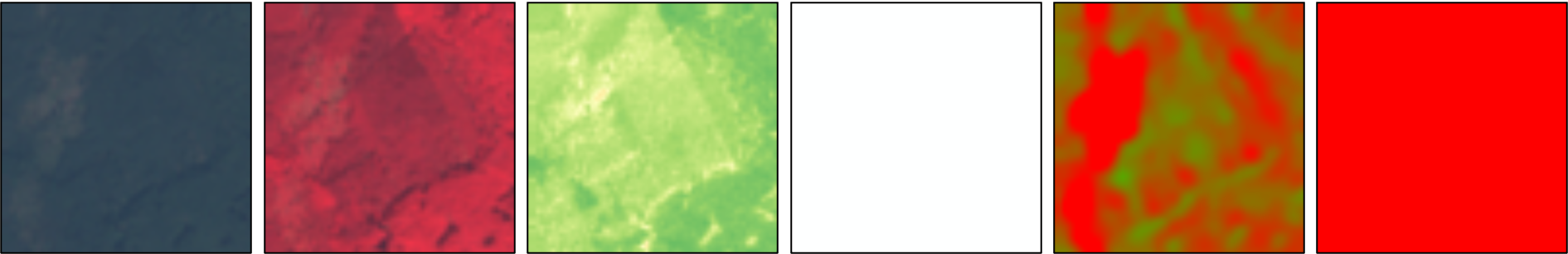}
  \vspace{-5mm}
    \caption{Example results on EuroSat. \textbf{Top:} Our multi-spectral model with Chain-of-Thought (CoT) reasoning correctly predicts `River', whereas the RGB-only baseline outputs `Highway'. The multi-spectral inputs, particularly the NDWI (4-th image), clearly distinguish water bodies where RGB features are ambiguous. \textbf{Bottom:} A `Forest' example correctly identified by our method. The RGB-only baseline misinterprets the green/blue features and incorrectly classifies the scene as `Sea lake'.
  }
  \label{fig:results_examples}
\end{figure}

\begin{table}[]
\caption{Ablation Study on BigEarthNet (Subset). We analyze the impact of input modalities across three prompting strategies: Baseline, Vocabulary Expansion, and Chain-of-Thought (CoT). While Expansion improves over the baseline, CoT yields robust gains even on RGB inputs, with the combination of CoT and all multi-spectral bands achieving the best performance.}
\vspace{-1mm}
\centering
\tablestyle{6.5pt}{1.1}
{
\begin{tabular}{l|l|c}
\toprule
{Prompting method} & {Input modality} & {F1} \\ 
\midrule
\multirow{4}{*}{\bf{Baseline (\ref{sec:generic})}} & RGB Only & 0.410 \\
 & RGB + NDVI & 0.448 \\
 & RGB + NDVI + NDWI & 0.442 \\
 & All Multi-Spectral & 0.479 \\ 
\midrule
\multirow{1}{*}{\bf{Expansion (\ref{sec:expansion})}} & All Multi-Spectral & 0.526 \\ 
\midrule
\multirow{4}{*}{\bf{CoT (\ref{sec:cot})}} & RGB Only & 0.529 \\
 & RGB + NDVI & 0.528  \\
 & RGB + NDVI + NDWI & 0.519  \\
 & \textbf{All Multi-Spectral} & \textbf{0.534} \\ 
\midrule
CoT w/o band description & All Multi-Spectral & 0.523 \\
CoT w/o pseudo-image description & All Multi-Spectral & 0.518 \\
\bottomrule
\end{tabular}%
\vspace{-2mm}
\label{tab:big_earth_ablations}
}
\end{table}

\textbf{Visualizations.}
Figure~\ref{fig:results_examples} illustrates instances where multi-spectral inputs are decisive. In the `River' example (top), the deep blue/green hues in the RGB image prove ambiguous, leading the RGB-only baseline to incorrectly predict `Highway'. Conversely, our proposed model leverages multi-spectral data and Chain-of-Thought reasoning to correctly identify the scene. Similarly, for the `Forest' example (bottom), the baseline misclassifies the input as `Sea lake' due to color confusion, whereas our method yields the correct classification.

\textbf{Limitations:} Our method relies on translating sensor data into visual inputs that generalist LMMs can interpret. While many types of aerial Earth observation data, e.g. multi-spectral, hyper-spectral, thermal, LiDAR, Radar, and SAR inputs, are suitable for the approach, some other sources of observation data may not necessarily map well to visual inputs.

\section{Conclusions}

We presented a training-free adaptation strategy to enable RGB-based LMMs to process multi-spectral Remote Sensing data. By converting spectral bands into instructional pseudo-images and applying Chain-of-Thought reasoning, we achieved new state-of-the-art Zero-Shot performance on BigEarthNet and EuroSat. This work highlights the latent potential of generalist models to solve specialized domain tasks when provided with interpretable, multi-modal context.

\small
\bibliographystyle{IEEEtranN}
\bibliography{references}

@String(PAMI = {IEEE Trans. Pattern Anal. Mach. Intell.})

@String(CVPR= {IEEE Conf. Comput. Vis. Pattern Recog.})

@String(ICCV= {Int. Conf. Comput. Vis.})

@String(ECCV= {Eur. Conf. Comput. Vis.})

@String(ICLR = {Int. Conf. Learn. Represent.})

@String(AAAI = {AAAI})

@String(CVPRW= {IEEE Conf. Comput. Vis. Pattern Recog. Worksh.})

@String(PAMI  = {IEEE TPAMI})

@String(CVPR  = {CVPR})

@String(ICCV  = {ICCV})

@String(ECCV  = {ECCV})

@String(ICLR  = {ICLR})

@String(CVPRW= {CVPRW})

@article{Llama3-MS-CLIP,
  title={Beyond the Visible: Multispectral
Vision-Language Learning for Earth Observation},
  author={Clive Tinashe Marimo and Benedikt Blumenstiel
and Maximilian Nitsche and Johannes Jakubik and Thomas Brunschwiler},
  journal={ECML PKDD},
  year={2025},
}

@article{GRAFT,
  title={Remote Sensing Vision-Language Foundation Models without Annotations via Ground Remote Alignment},
  author={Utkarsh Mall and Cheng Perng Phoo and Meilin Kelsey Liu and Carl Vondrick and Bharath Hariharan and Kavita Bala},
  journal=iclr,
  year={2024},
}

@article{BigEarthNet19Cl,
  title={BigEarthNet-MM: A Large Scale Multi-Modal Multi-Label Benchmark Archive for Remote Sensing Image Classification and Retrieval},
  author={Gencer Sumbul and Arne de Wall and Tristan Kreuziger and Filipe Marcelino and Hugo Costa and Pedro Benevides and Mário Caetano and Begüm Demir and Volker Markl},
  journal={IEEE Geoscience and Remote Sensing Magazine},
  year={2021},
}

@article{cot,
  title={Chain-of-Thought Prompting Elicits Reasoning
in Large Language Models},
  author={Jason Wei Xuezhi Wang Dale Schuurmans Maarten Bosma
Brian Ichter Fei Xia Ed H. Chi Quoc V. Le Denny Zhou},
  journal=neurips,
  year={2022},
}

@article{MMEarth,
  title={MMEarth: Exploring Multi-Modal Pretext Tasks For Geospatial Representation Learning},
  author={Vishal Nedungadi and Ankit Kariryaa  and Stefan Oehmcke and Serge Belongie and
Christian Igel and Nico Lang},
  journal=eccv,
  year={2024},
}

@article{OmniSat,
  title={OmniSat: Self-Supervised Modality Fusion for Earth Observation},
  author={Guillaume Astruc and Nicolas Gonthier and Clement Mallet and Loic Landrieu},
  journal=eccv,
  year={2024},
}

@article{CROMA,
  title={CROMA: Remote Sensing Representations with Contrastive Radar-Optical Masked Autoencoders},
  author={Anthony Fuller and Koreen Millard and James R. Green},
  journal=neurips,
  year={2023},
}

@article{Billion,
  title={A Billion-scale Foundation Model for Remote Sensing Images},
  author={Keumgang Cha, Junghoon Seo, Taekyung Lee},
  journal={IEEE Journal of Selected Topics in Applied Earth Observations and Remote Sensing (IEEE J-STARS)},
  year={2023},
}

@article{GFM,
  title={Towards Geospatial Foundation Models via Continual Pretraining},
  author={Matias Mendieta and Boran Han and Xingjian Shi and Yi Zhu and Chen Chen},
  journal=iccv,
  year={2023},
}

@article{SpectralSpatial,
  author={Ibañez, Damian and Fernandez-Beltran, Ruben and Pla, Filiberto and Yokoya, Naoto},
  journal={IEEE Transactions on Geoscience and Remote Sensing}, 
  title={Masked Auto-Encoding Spectral–Spatial Transformer for Hyperspectral Image Classification}, 
  year={2022},
  doi={10.1109/TGRS.2022.3217892}
  }

@article{MaskedHyperspectral,
  title={Masked Vision Transformers for Hyperspectral Image Classification},
  author={Linus Scheibenreif and Michael Mommert and Damian Borth},
  journal={EEE/CVF Conference on Computer Vision and Pattern Recognition Workshops (CVPRW)},
  year={2023},
}

@article{hong2022spectralformer,
  title={Spectralformer: Rethinking hyperspectral image classification with transformers},
  author={Hong, Danfeng and Han, Zhu and Yao, Jing and Gao, Lianru and Zhang, Bing and Plaza, Antonio and Chanussot, Jocelyn},
  journal={IEEE Trans. Geosci. Remote Sens.},
  year={2022},
  volume={60},
  pages={1-15},
  note = {DOI: 10.1109/TGRS.2021.3130716}
}

@article{Scale-MAE,
  title={Scale-MAE: A Scale-Aware Masked Autoencoder for Multiscale Geospatial
Representation Learning},
  author={Colorado J Reed and Ritwik Gupta and Shufan Li and
 Sarah Brockman and
 Christopher Funk and
 Brian Clipp and
Kurt Keutzer and Salvatore Candido and Matt Uyttendaele and Trevor Darrell},
  journal=iccv,
  year={2023},
}

@article{BigEarthNet,
  title={BigEarthNet: A Large-Scale Benchmark Archive For Remote Sensing Image Understanding},
  author={Gencer Sumbul and Marcela Charfuelan and Begüm Demir and Volker Markl},
  journal={IEEE International Geoscience and Remote Sensing Symposium (IGARSS)},
  year={2019},
  publisher={IEEE}
}

@article{helber2019eurosat,
  title={Eurosat: A novel dataset and deep learning benchmark for land use and land cover classification},
  author={Helber, Patrick and Bischke, Benjamin and Dengel, Andreas and Borth, Damian},
  journal={IEEE Journal of Selected Topics in Applied Earth Observations and Remote Sensing},
  year={2019},
  publisher={IEEE}
}

@inproceedings{kuckreja2023geochat,
          title={GeoChat: Grounded Large Vision-Language Model for Remote Sensing},
          author={Kuckreja, Kartik and Danish, Muhammad S. and Naseer, Muzammal and Das, Abhijit and Khan, Salman and Khan, Fahad S.},
          journal=cvpr,
          year={2024}
  }

@inproceedings{LoRA-NIR,
    title={LoRA-NIR: Low-Rank Adaptation of Vision Transformers for Remote Sensing With Near-Infrared Imagery},
    author={Irem Ulku and O. Ozgur Tanriover and Erdem Akagündüz},
    booktitle={IEEE Geoscience and Remote Sensing Letters},
    year={2024},
}

@inproceedings{Gemini2.5,
    title={Gemini 2.5: Pushing the Frontier with Advanced Reasoning, Multimodality, Long Context, and Next Generation Agentic Capabilities},
    author={Gemini Team},
    booktitle={arxiv.org/abs/2507.06261},
    year={2025},
}

@inproceedings{s2vec,
    title={S2Vec: Self-Supervised Geospatial Embeddings},
    author={Shushman Choudhury and Elad Aharoni and Chandrakumari Suvarna and Iveel Tsogsuren and Abdul Rahman Kreidieh and Chun-Ta Lu and Neha Arora},
    booktitle={https://arxiv.org/abs/2504.16942},
    year={2025},
}

@inproceedings{SatlasPretrain,
    title={SatlasPretrain: A Large-Scale Dataset for Remote Sensing Image Understanding},
    author={Favyen Bastani and Piper Wolters and Ritwik Gupta and Joe Ferdinando and Aniruddha Kembhavi},
    booktitle=iccv,
    year={2023},
}

@inproceedings{MOSAIKS,
    title={A generalizable and accessible approach to machine learning with global satellite imagery},
    author={Esther Rolf and Jonathan Proctor and Tamma Carleton and Ian Bolliger and Vaishaal Shankar and Miyabi Ishihara and Benjamin Recht and Solomon Hsiang},
    booktitle={arxiv.org/abs/2010.08168},
    year={2020},
}

@inproceedings{AlphaEarth,
    title={AlphaEarth {F}oundations: An embedding field model for accurate and efficient global
mapping from sparse label data},
    author={Christopher F. Brown and Michal R. Kazmierski and Valerie J. Pasquarella and William J. Rucklidge and Masha Samsikova and Chenhui Zhang and Evan Shelhamer and Estefania Lahera and Olivia Wiles and Simon Ilyushchenko and Noel Gorelick and Lihui Lydia Zhang and Sophia Alj and Emily Schechter and Sean Askay and Oliver Guinan and Rebecca Moore and Alexis Boukouvalas and Pushmeet Kohli},
    booktitle={arxiv.org/pdf/2507.22291 },
    year={2025},
}

@inproceedings{ZLaP,
    title={Label Propagation for Zero-shot Classification with Vision-Language Models},
    author={Vladan Stojnic and Yannis Kalantidis and Giorgos Tolias},
    booktitle=cvpr,
    year={2024},
}

@inproceedings{EarthPT,
    title={EarthPT: a time series foundation model for Earth Observation
},
    author={Michael J. Smith and Luke Fleming and James E. Geach},
    booktitle={arxiv:2309.07207},
    year={2023},
}

@inproceedings{HyperSigma,
    title={HyperSIGMA: Hyperspectral Intelligence Comprehension Foundation Model},
    author={Di Wang and Meiqi Hu and Yao Jin and Yuchun Miao and Jiaqi Yang and Yichu Xu and Xiaolei Qin and Jiaqi Ma and Lingyu Sun and Chenxing Li and Chuan Fu and Hongruixuan Chen and Chengxi Han and Naoto Yokoya and Jing Zhang and Minqiang Xu and Lin Liu and Lefei Zhang and Chen Wu and Bo Du and Dacheng Tao and Liangpei Zhang},
    booktitle=PAMI,
    year={2025},
}

@inproceedings{SpectralGPT,
    title={SpectralGPT: Spectral Remote Sensing Foundation Model},
    author={Danfeng Hong and Bing Zhang and Xuyang Li and Yuxuan Li and Chenyu Li and Jing Yao and Naoto Yokoya and Hao Li and Pedram Ghamisi and Xiuping Jia and Antonio Plaza and Paolo Gamba and Jon Atli Benediktsson and Jocelyn Chanussot},
    booktitle=PAMI,
    year={2024},
}

@inproceedings{FoundSurvey,
    title={Vision Foundation Models in Remote Sensing: A Survey},
    author={Siqi Lu and Junlin Guo and James R Zimmer-Dauphinee and Jordan M Nieusma and Xiao Wang and Parker VanValkenburgh and Steven A Wernke and Yuankai Huo},
    booktitle={arxiv:2408.03464},
    year={2024},
}

@inproceedings{Interband,
    title={Interband Retrieval and Classification Using the Multilabeled Sentinel-2 BigEarthNet Archive},
    author={Ushasi Chaudhuri and Subhadip Dey and Mihai Datcu and Biplab Banerjee and Avik Bhattacharya},
    booktitle={IEEE Journal of Selected Topics in Applied Earth Observations and Remote Sensing},
    year={2021},
}

@inproceedings{linial2025Enhancing,
    title={Enhancing Remote Sensing Representations Through Mixed-Modality Masked Autoencoding},
    author={Ori Linial and George Leifman and Yochai Blau and Nadav Sherman and Yotam Gigi and Wojciech Sirko and Genady Beryozkin},
    booktitle={ Winter Conference on Applications of Computer Vision (WACV) Workshops},
    year={2025},
}

@article{Prithvi-100M-preprint,
    author          = {Jakubik, Johannes and Roy, Sujit and Phillips, C. E. and Fraccaro, Paolo and Godwin, Denys and Zadrozny, Bianca and Szwarcman, Daniela and Gomes, Carlos and Nyirjesy, Gabby and Edwards, Blair and Kimura, Daiki and Simumba, Naomi and Chu, Linsong and Mukkavilli, S. Karthik and Lambhate, Devyani and Das, Kamal and Bangalore, Ranjini and Oliveira, Dario and Muszynski, Michal and Ankur, Kumar and Ramasubramanian, Muthukumaran and Gurung, Iksha and Khallaghi, Sam and Li, Hanxi (Steve) and Cecil, Michael and Ahmadi, Maryam and Kordi, Fatemeh and Alemohammad, Hamed and Maskey, Manil and Ganti, Raghu and Weldemariam, Kommy and Ramachandran, Rahul},
    month           = oct,
    title           = {{Foundation Models for Generalist Geospatial Artificial Intelligence}},
    journal         = {Preprint Available on arxiv:2310.18660},
    year            = {2023}
}

@inproceedings{SkySense,
    title={SkySense: A Multi-Modal Remote Sensing Foundation Model Towards Universal Interpretation for Earth Observation Imagery},
    author={Xin Guo and Jiangwei Lao and Bo Dang and Yingying Zhang and Lei Yu and Lixiang Ru and Liheng Zhong and Ziyuan Huang and Kang Wu and Dingxiang Hu and Huimei He and Jian Wang and Jingdong Chen and Ming Yang and Yongjun Zhang and Yansheng Li},
    booktitle=cvpr,
    year={2024},
}

@inproceedings{RemoteCLIP,
    title={Remote{C}{L}{I}{P}: A Vision Language Foundation
Model for Remote Sensing},
    author={Fan Liu and Delong Chen and Zhangqingyun Guan and Xiaocong Zhou and Jiale Zhu and Qiaolin Ye and Liyong Fu and Jun Zhou},
    booktitle={IEEE Transactions on Geoscience and Remote Sensing (TGRS)},
    year={2024},
}

@inproceedings{skyscript,
    title={SkyScript: A Large and Semantically Diverse Vision-Language Dataset for Remote Sensing},
    author={Zhecheng Wang and Rajanie Prabha1 and Tianyuan Huang and Jiajun Wu and Ram Rajagopal},
    booktitle=AAAI,
    year={2024},
    url={https://arxiv.org/pdf/2312.12856v1}
}

@inproceedings{satmae2022,
    title={Sat{MAE}: Pre-training Transformers for Temporal and Multi-Spectral Satellite Imagery},
    author={Yezhen Cong and Samar Khanna and Chenlin Meng and Patrick Liu and Erik Rozi and Yutong He and Marshall Burke and David B. Lobell and Stefano Ermon},
    booktitle={Advances in Neural Information Processing Systems},
    editor={Alice H. Oh and Alekh Agarwal and Danielle Belgrave and Kyunghyun Cho},
    year={2022},
    url={https://arxiv.org/abs/2207.08051}
}

@inproceedings{RSCLIP,
    title={RS-CLIP: Zero shot remote sensing scene classification via contrastive vision-language supervision},
    author={Xiang Li and Congcong Wen and Yuan Hu  and Nan Zhou},
    booktitle={International Journal of Applied Earth Observation and Geoinformation},
    year={2023},
    url={https://www.sciencedirect.com/science/article/pii/S1569843223003217}
}

@inproceedings{Zhang2024GoodAt,
       title={Good at captioning, bad at counting: Benchmarking GPT-4V on Earth observation data},
	    author={Chenhui Zhang and Sherrie Wang},
	    booktitle={arxiv.org/pdf/2401.17600},
	    year={2024}
    }

@article{mallya2025zero,
  title={Zero-Shot Multi-Spectral Learning: Reimagining a Generalist Multimodal Gemini 2.5 Model for Remote Sensing Applications},
  author={Mallya, Ganesh and Gigi, Yotam and Kim, Dahun and Neumann, Maxim and Beryozkin, Genady and Shekel, Tomer and Angelova, Anelia},
  journal={arXiv preprint arXiv:2509.19087},
  year={2025}
}

@inproceedings{radford2021clip,
      title={Learning Transferable Visual Models From Natural Language Supervision}, 
      author={Alec Radford and Jong Wook Kim and Chris Hallacy and Aditya Ramesh and Gabriel Goh and Sandhini Agarwal and Girish Sastry and Amanda Askell and Pamela Mishkin and Jack Clark and Gretchen Krueger and Ilya Sutskever},
      booktitle={ICML},
      year={2021},
}

@article{Barzilai2025ARF,
  title={A Recipe for Improving Remote Sensing VLM Zero Shot Generalization},
  author={Aviad Barzilai and Yotam Gigi and Vered Silverman and Yehonathan Refael and Bolous Jaber and Amr Helmy and Tomer Shekel and George Leifman and Genady Beryozkin},
  journal={ArXiv},
  year={2025},
  volume={abs/2503.08722},
  url={https://api.semanticscholar.org/CorpusID:276937917}
}

\end{document}